\documentclass[journal]{journal}
\usepackage{graphicx}
\ifCLASSINFOpdf
\else
\fi
\begin{document}
%
\title{Individual risk profiling for portable devices using a neural network to process the cognitive reactions and the emotional responses to a multivariate situational risk assessment}
%
%
%

\author{Frederic Jumelle, 
        Kelvin So, 
        and Didan Deng,
\thanks{Dr. Frederic Jumelle is with Bright Nation Limited, Smart-Space 3F, Cyberport, Hong Kong. E-mail: (see f.jumelle@brightnationlimited.com).}
\thanks{Kelvin So is with Bright Nation Limited, Smart-Space 3F, Cyberport, Hong Kong. E-mail: (see ccso@brightnationlimited.com).}
\thanks{Didan Deng is with Neuromorphic Interactive System Laboratory, Department of Electronic and Computer Engineering, The Hong Kong University of Science and Technology, Hong Kong. E-mail: (ddeng@ust.hk).}}

%
%

\markboth{Journal of \LaTeX\ Class Files,~Vol.~6, No.~1, January~2007}%
{Shell \MakeLowercase{\textit{et al.}}: Bare Demo of IEEEtran.cls for Journals}
%



\maketitle
\thispagestyle{empty}

\begin{abstract}
In this paper, we are presenting a novel method and system for neuropsychological performance testing that can establish a link between cognition and emotion. It comprises a portable device used to interact with a cloud service which stores user information under username and is logged into by the user through the portable device; the user information is directly captured through the device and is processed by artificial neural network; and this tridimensional information comprises user cognitive reactions, user emotional responses and user chronometrics. The multivariate situational risk assessment is used to evaluate the performance of the subject by capturing the 3 dimensions of each reaction to a series of 30 dichotomous questions describing various situations of daily life and challenging the user’s knowledge, values, ethics, and principles. In industrial application, the timing of this assessment will depend on the user’s need to obtain a service from a provider such as opening a bank account, getting a mortgage or an insurance policy, authenticating clearance at work or securing online payments.
\end{abstract}


%
\IEEEpeerreviewmaketitle

\section{Introduction}
Based on the theory of cognitive neuroscience and experimental neuropsychology, neurolinguistics and computer models, the relationship between the psychological phenomenon of the subject and the brain structure has been established. The investigation techniques based on experimental neuroscience include functional magnetic resonance imaging, positron emission tomography scanner, electroencephalography, and magnetoencephalography. Applied neuroscience has been integrating research results from different fields and on different scales enough to reach a unified descriptive model of the brain functioning and biosocial expression of the personality. 

With the development of cognitive neuroscience, Professor Robert Cloninger proposed a unified biosocial theory of personality~\cite{cloninger1986unified} and a systematic method~\cite{cloninger1987systematic} based on a multidimensional personality questionnaire~\cite{cloninger1991tridimensional} to assess different facets of the personality: the Temperament and Character Inventory (TCI)~\cite{cloninger1994temperament}. He believes that the method of obtaining accurate neuropsychological performance~\cite{cloninger1993psychobiological} not only needs to consider behavioral factors, but also needs to consider potential biological and social determinants - and distinguish between the perceptual and the conceptual factors. TCI aims to distinguish the hereditary nature (Temperament) from the acquired nature (Character) of the personality development~\cite{cloninger1997role} through an experimental method (Inventory) to obtain the subject’s neuropsychological performance. TCI can also be used to identify various personality disorders~\cite{cloninger1997integrative} to examine the extent of personality disorder development and the relationships between personality and depression~\cite{corruble2002early, chen2011effects} or between personality and alcohol dependence~\cite{bjornebekk2012social} and between personality and crime~\cite{maristemperament} and serve as a model using AI to predict patient’s relapse~\cite{maldonato2018non}. It was also used to approach non-clinical population~\cite{cloninger2004feeling, jumelle2007temperament} consumer habits~\cite{pozharliev2017social} and the mechanism behind choices~\cite{sperandeo2016executive} especially the role of the executive functions. 

Traditional TCI-R approach is more about auto-assessment rather than investigation which makes it a relatively biased intervention that obviously misses the emotional assessment and the relationship between cognitive stimulation and emotional response.

In this paper, we are proposing a neuropsychological performance test for the subject that is not only assessing the temperament of the subject but also the correlation between the subject's emotional response to a series of cognitive questions regarding situational risk. The answers to the questions constitute the risk tolerance of the subject while the degree of correlation between the emotional response and the cognitive questions is an indicator of the worthiness of the subject. This motivates us to develop an accurate and efficient machine learning framework for risk profiling and creditworthiness indicator which takes both edges from traditional neuroscience techniques, for example, TCI, and the recent advancement of neuromorphic intelligence design~\cite{wang2015formal} and neural network models especially those dedicated to facial emotion recognition by facial muscles movement analysis~\cite{deng2019mimamo}. 

In our artificial neural network model for risk profiling, we are processing simultaneously the emotion response captured by a prominent emotion recognition model, MIMAMO-Net~\cite{deng2019mimamo}, with the cognitive response to a question describing a situational risk also described as a choice between 2 consequential often opposite behaviors when submitted to a situation involving a prescribed level of risk. The subject has to make a Yes/No choice (answer) to a question describing a risky situation and this Q\&A is repeated 30 times. It involves 30 situations of various risky nature to cover the daily life topics and interactions of an average adult individual. Each user receives a unique selection of 30 questions from the local database of 1,200 questions. The 3 main dimensions of temperament HA, RD and NS are equally represented and assembled in pairs of binomial nature such as HA/RD, HA/NS, RD/HA, RD/NS, NS/HA and NS/RD and repeatedly proposed during the 30 items long questionnaire. During the questionnaire, the subject facial reaction to the questions or emotional response will be captured continuously and the reaction time between questions and answers will be recorded.

The video capture of the facial gesture of the subject when answering each question is recorded and then fed-in to MIMAMO-Net to generate a arousal-valence score for each question answered.

Our neural network can open a new perspective to the non-invasive field of investigation of the hidden processes of the human brain. Using AI to process simultaneously a double stream of recorded data under time pressure, emotional responses and cognitive reactions, and compare them in real-time has the interest to remove the many biases of current techniques. Because the questionnaire is operated by a mobile application that allows flexibility on the time of the first investigation and enables further use of the results especially in the form of an encrypted QR Code, it may represent a supplementary form of personal identity based on personal attributes to be vectorized by the user for policy-based authentication, login and payment.

\section{Methodology}

\subsection{Situational Risk Tolerance Assessment (SRTA)}

Our SRTA is a method and system for neuropsychological performance test inspired by the TCI-R of Cloninger. The SRTA is used to evaluate the reactional performance of the subject to a series of dichotomous questions describing various situations of daily life and challenging the subject values, ethics, principles or virtues. The performance depends on a combination of personal characteristics: the biological characteristics of the subject such as physical health factors, genetic vulnerability, addictive behaviors; the social characteristics such as family environment, close relationships, marital status; and the psychological characteristics such as cooperative skills, social skills, relational skills, self-esteem and mental health. During the industrial application of the SRTA, the time of assessment also depends on the subject's need to obtain a service from a provider such as opening an account in a bank, getting a quote of an insurance premium or an authentication key for a job with clearance or securing online payments. 

\subsection{Neural network model for risk profiling (ANN)} \label{ANN_s}

\begin{figure*}[t]
\centering
\centerline{\includegraphics[height=1.8in]{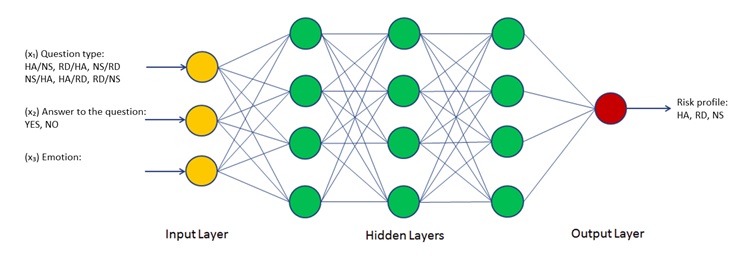}}
\caption{The architecture of our neural network (ANN)}\label{ANN}
\end{figure*}

We build our neural network from the standard building blocks of fully connected layers~\cite{hassoun1995fundamentals, lecun2015deep, wasserman1993advanced, white1992artificial}. To put it in a mathematical manner, our data can be expressed as
$$
\left\{ \left( x_1, x_2, x_3 \right)_n \right\}_{n = 1, \ldots, N}
$$
where $x_1$ is the question type (HA/NS, RD/HA, NS/RD, NS/HA, HA/RD, RD/NS), $x_2$ is the answer to question (Yes/No), $x_3$ is the emotional states (valence and arousal) of the subject and $N$ is the number of questions answered (and thus the number of emotional states recorded). The output is a risk profile and take value either HA, RD or NS. Figure~\ref{ANN} is a schematic diagram of the feed forward artificial neural network, also called a multilayer perceptron, of the neuropsychological performance testing system proposed in this paper. It is a supervised, fully connected, feed-forward artificial neural network and uses back-propagation training and generalized delta rule learning~\cite{khalil2009performance, taner1995neural}.

\paragraph{Major question types - labeling with cognitive tags for machine learning} The questions are separated in two groups, namely, the major types and the minor types. The major types are related to straightforward and obvious situations or issues in life while the minor types are related to more complex ideas and subjective problems. In our questionnaire, the major type questions appear few times more frequent than the minor types. For example, the following questions are major types: 

\begin{itemize}
\item Do you like gardening? HA/NS 

\item Do you like dancing? RD/HA 

\item Do you seek revenge after being hurt? NS/RD 
\end{itemize}

Each question has its specific bi-dimensional properties. The first question: "Do you like gardening?" is a major type of question related to Harm Avoidance (HA) versus Novelty Seeking (NS). When the answer to this question is Yes, it means that the temperament of the test subject or user is in favor of HA. HA refers to harm avoidance also called risk aversion (Averse). To the contrary if the answer is No, the temperament would be in favor of NS. NS refers to novelty seeking also called risk seeking (Taker). A similar approach is applied to a RD/HA question type such as "Do you like dancing?". If the user chooses to answer No, it would in favor of HA which is for risk aversion (Averse); if the user's response is Yes, it would be in favor of RD which refers to Reward Dependence (RD) also called risk dependent (Dependent). 

\paragraph{Minor question types - labeling with cognitive tags for machine learning} The present invention questionnaire is also using a set of minor type questions in lesser number of occurrences, such as: 

\begin{itemize}
\item Do you feel happy while spending money? NS/HA 
\item Do you keep your problems to yourself? HA/RD 
\item Do you consider right from wrong before taking decisions? RD/NS 
\end{itemize}

Alike the major types of question, each minor question has its own bi-dimensional properties referring to temperament characteristics. The exact same processing of information of the major types applies to the minor types.

Each answer to a question is granting one data-point in one of the 3 dimensions/bins and the highest score of the 3 bins where the most data-points are accumulated will determine the first trait of temperament of the user also called primary score of the risk profile. The second bin with the second highest number of data-points will determine the secondary score of the risk profile and the combination of primary and secondary scores will become the user risk profile score. 

The user brain processing time in reaction to the situational risk questions of the questionnaire is called latency and the performance test average latency information is the average time of the questionnaire test divided by the total number of questions asked. The range of the test average latency is theoretically between 0.2 and 10 seconds. A reaction time of 0.2 second is considered to be the minimum of reactional time in the human benchmark that doesn’t include any reflection or processing time of the question. A maximum processing time of 10 seconds is preset to avoid stalling. The pilot population has showed an average latency between 2 and 7 seconds per question.  The result obtained by a test user is sorted and put into a five categories model. 

In the five categories model, the test population is divided into extra short (XS), short, median, long, extra long (XL) corresponding to 3 standard deviations. A brain processing time or latency of 0.5 to 1.5 seconds is regarded as extra short; 1.5 to 2.5 seconds is short; 2.5 to 3.5 seconds is median; 3.5 to 4.5 seconds is long; 4.5 to 7 seconds is extra long. The median of population is around 3 seconds. Any result below 2 or above 7 seconds is considered unusual. More details on the mathematical definition of the five categories can be found in section~\ref{IWI} and figure~\ref{normal}. The user can skip a question if he doesn’t understand or like it and a new question of similar type is called, this event is called a revalidation. Only 6 revalidations are allowed.

If there are more than 6 revalidations, the whole questionnaire result is deemed untruthful therefore invalid. While the questionnaire is being activated, the camera is starting to capture the video of the survey which is divided in 2 parts: the capture of the facial emotion information immediately after the reception of the question on the screen also called 'reaction time' for which the minimum is approximately 0.2 second - and processed by the mobile AI for facial recognition; then the video capture continues during the reflection time until the response is decided by the test user whether by pushing the Yes or No button on the screen or by recording a vocal answer or both text and voice if voice-movement coordination patterns recognition is activated. These video captures are fed into MIMAMO-Net for emotion recognition, with an output of valence-arousal pair.

The facial expression information or emotional expressions of the test subject are compared with pre-stored biometric information, also called training data. For the matched information, it will determine whether the input is normal and the biometric type is corresponding to the expected emotion of the test subject; or if it deviates, how much is the deviation, and what is the deviation that would be sufficient to disqualify the record. If the mobile AI of the invention disqualifies an emotion to a question, a new question of the similar type will be asked at the end of the 30 questions standard model and will make the questionnaire of 31 questions, 30 standard plus 1 revalidation that will contribute to assess the truthfulness. 



\paragraph{Training process} Backpropagation is the core part of the training and consists of two phases: stimulus propagation and weight update. 

The excitation propagation phase consists of two steps: \textbf{1.} Forward propagation: the training input is sent to the network to obtain the excitation response; \textbf{2.} Back propagation: the excitation response corresponds to the training input. The target output is evaluated to obtain a response error of the output layer and the hidden layer. 

In the weight update phase, update is made for each synapse weight in two steps: \textbf{1.} Multiply the input stimulus and response error to obtain a gradient of weights; \textbf{2.} Multiply this gradient by a scale and invert it added to the weight; this ratio (percentage) will affect the speed and effect of the training process, thus becoming a "training factor."

\begin{figure*}[t]
\centering
\centerline{\includegraphics[height=1.8in]{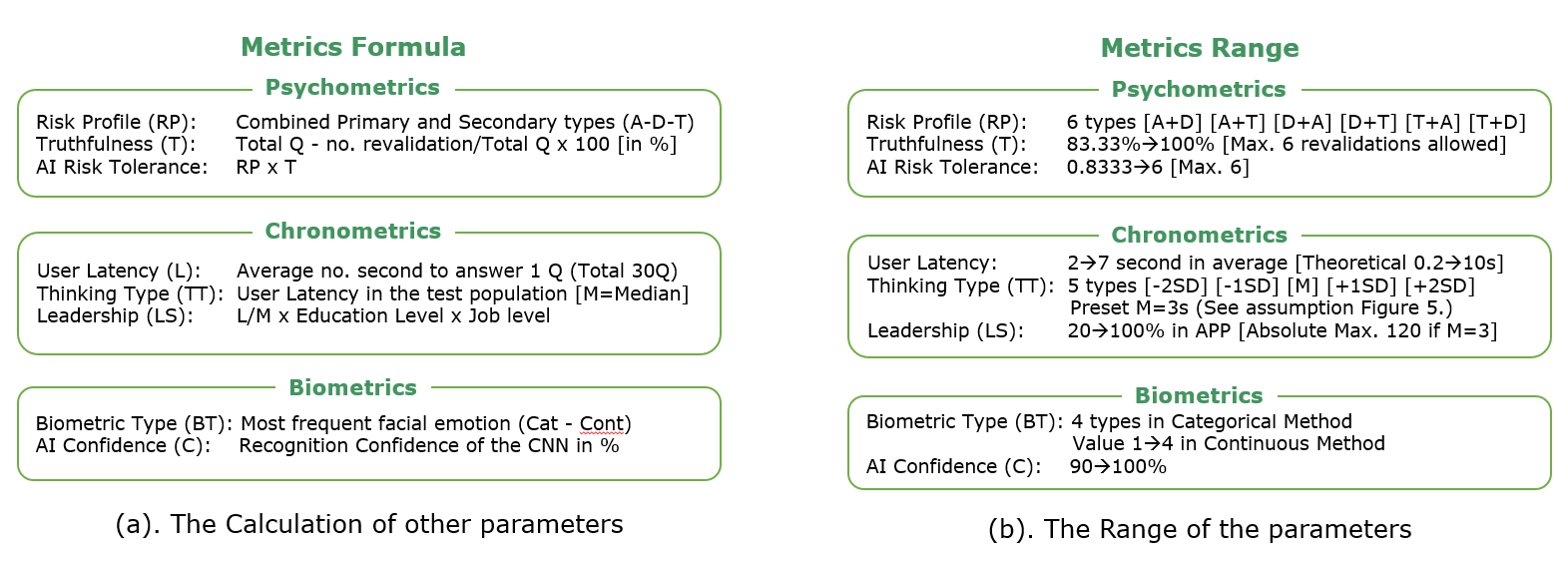}}
\caption{The calculation and range of parameters}\label{other_parameters}
\end{figure*}

\begin{figure*}[t]
\centering
\centerline{\includegraphics[height=1.8in]{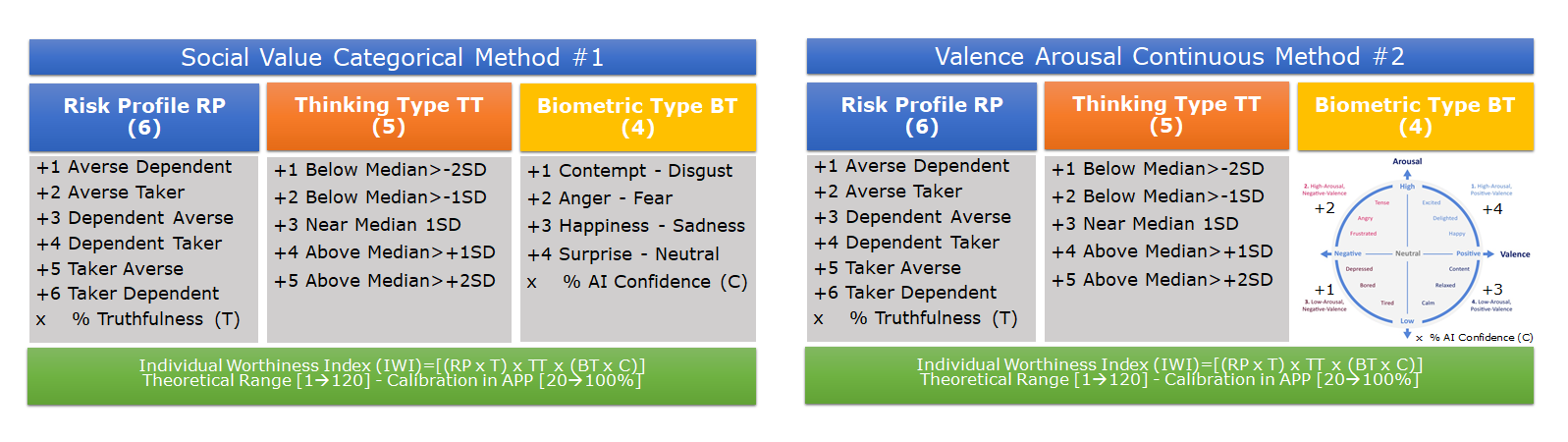}}
\caption{The calculation of the Individual Worthiness Index (IWI) using Categorical and Continuous Method}\label{IWI1}
\end{figure*}

The two phases iterate until the network's response to the input reaches a satisfactory predetermined target range. In the below, we will enumerate the training steps with reference to figure~\ref{ANN}.

\begin{enumerate}
\item Note that the output of each node is given by: 
$$ y = \bar{w} x  + \bar{b} $$
where $\bar{w}$ is the weight and $\bar{b}$ is the bias.
The output layer is: 
$$ \phi \left( \bar{w}_0 \cdot \left( \sum \bar{w}_h x + \bar{b}_h \right) + \bar{b}_0 \right) $$
where $\bar{w}_h$ and $\bar{b}_h$ are the weight and bias of the last hidden layer, and $\bar{w}_0$ and $\bar{b}_0$ are the weight and bias of the output layer and the activation function $\phi$ is ReLU~\cite{agarap2018deep}.

\item
Backpropagation is then run to minimize the validation error: 
$$ \delta_h =  y_0 (1-y_0)(y_0-t) $$
where $y_0$ is the output of the output layer and $t$ is the target output (ground truth).

\item
For each node, the back-propagation error term is: 
$$ \delta_h' = y_h' (1-y_h') \sum \delta_h \cdot \bar{w}_h$$

\item
We will then update the synaptic weights from a node in layer $n$ to a node in layer $n+1$, 
$$ \bar{w} =: \bar{w} + \Delta \bar{w}$$
where $ \Delta \bar{w} = -y \delta_h' x$ and $y$ is the learning rate.

\item
Lastly, we calculate Mean Squared Errors:
$$ \textrm{MSE} = \frac{1}{N_{ts}} \cdot \sum_{1}^{N_{ts}} (y_0 - t)^2 $$ 

\end{enumerate} 

We will repeat the forward propagation and back propagation until the number of the epoch limit or early stopping criteria is reached. After training, the model runs on 15\% test dataset to calculate precision, accuracy and F score. 

\subsection{Individual Worthiness Index (IWI)} \label{IWI}

After initializing and training the neural network for risk profiling, the output from the neural network, e.g. risk profile, as well as other parameters will enable the delivery of the Individual Worthiness Index (IWI) to the subject.

\paragraph{Other parameters} Part (a) of figure~\ref{other_parameters} is an illustration of the various parameters of the present invention. It includes the following parameters:
\begin{itemize}
\item Risk Profile (RP)
\item The Truthfulness (T)
\item The AI Risk Tolerance (AIRT) is the score resulting of RP multiplied by T.
\item The User Latency (L)
\item The Thinking Type (TT)
\item The Leadership (LS) score depends on the ratio of user Latency over Median multiplied by Education and Job levels.
\item The Biometric Type (BT)
\item The AI Confidence score (C) is calculated by the mobile AI processing the facial emotion recognition at the front end of the invention. 
\end{itemize}

\paragraph{Range of the parameters} Part (b) of figure~\ref{other_parameters} is a table showing the ranges of the scores delivered at the end of processing by the present invention. Notice that 3 types of Metrics are presented and correlated to build a tri-dimensional 3D profile of the test subject or user. The first part is called Psychometrics and refers to the cognitive part of the performance test based on semantics. 

\begin{itemize}
\item The primary and secondary scores of the Risk Profile are given by the distribution of the answer’s datapoints during the survey in the 3 dimensions/bins.
\item The Risk Profile has 6 types resulting of 6 combinations of the 3 scores in pairs.
\item The Truthfulness is determined by the number of revalidation.
\item The truthfulness range is 83.33-100\%.
\end{itemize}

The second part is called Chrono-metrics and refers to the timing part (using a chronometer) of the performance test. It includes the following parameters and corresponding ranges:

\begin{itemize}
\item The average user latency of a full test is estimated between 2,000 and 7,000 milliseconds.
\item The Thinking Type has 5 types.
\item The Leadership score ranges between 0 and 120 when 6 levels of Education and Job are preset.
\end{itemize}

The third part is called the Biometrics and refers to the result of facial emotion recognition part of the performance test.

\begin{itemize}
\item The Biometric Type has 4 types.
\item The Confidence score in \% is a score delivered by the mobile AI regarding its ability to recognize the facial emotion accurately.
\end{itemize} 


\paragraph{The Individual Worthiness Index} Figure~\ref{IWI1} are tables dedicated to explanatory analysis using the performance test results of our invention whether combined with Categorical Method or Continuous Method. The test results are divided into three parts as in part (a) of figure~\ref{other_parameters}.

\begin{figure*}[t]
\centering
\centerline{\includegraphics[height=2.5in]{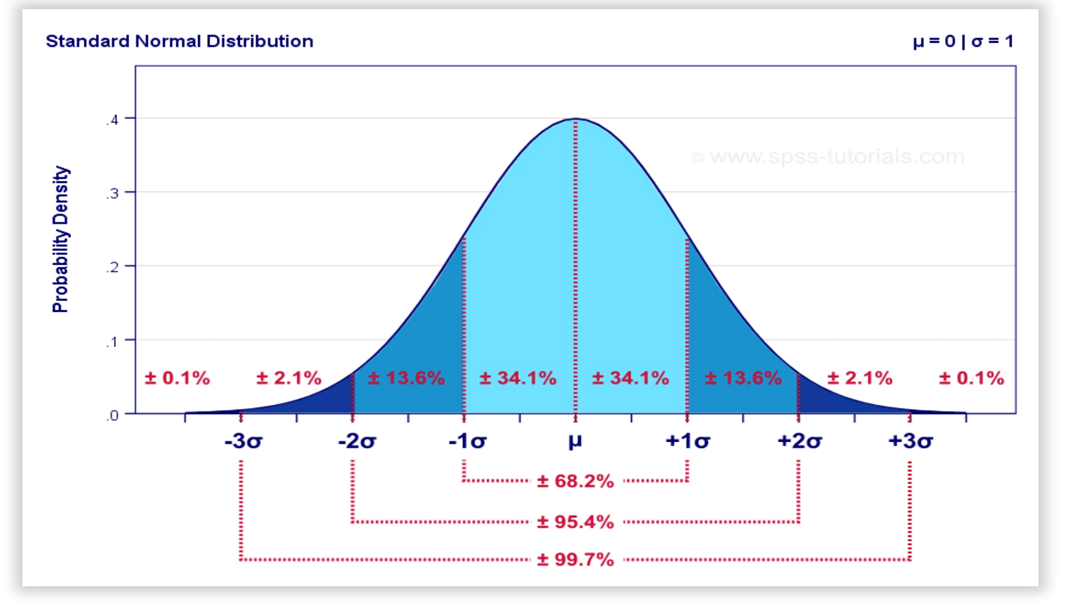}}
\caption{Assumption of distribution of latency for the Calculation of the Thinking Type (TT)}\label{normal}
\end{figure*}

\begin{figure*}[t]
	\centering
	\centerline{\includegraphics[height=1.8in]{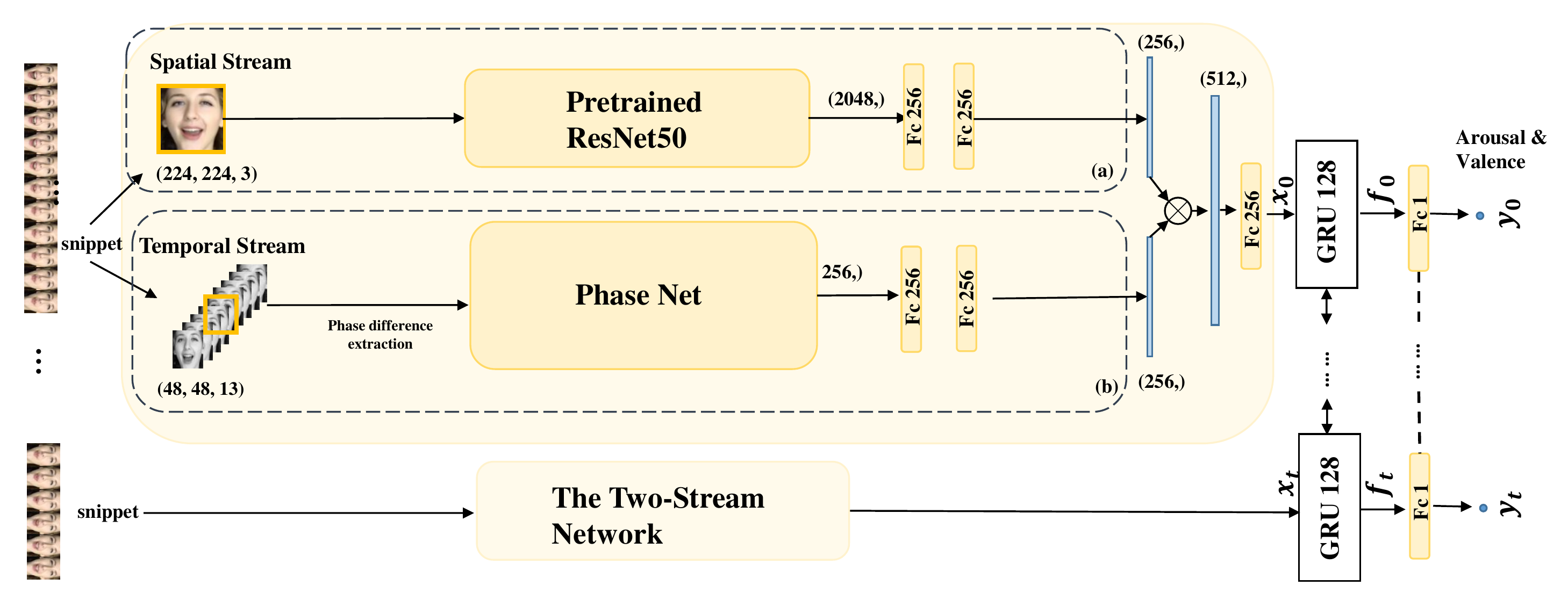}}
	\caption{The architecture of the MIMAMO Net}\label{mimamo-net}
\end{figure*}

%

The first part is the Risk Profile (RP) which divided into 6 categories that have been attributed a coefficient according to the social valuation of the risk scores combination (primary + secondary). For example, +1 for Averse Dependent; +2 for Averse Taker; +3 for Dependent Averse; +4 for Dependent Taker; +5 for Taker Averse; +6 for Taker Dependent. 

The second part is the Thinking Type (TT) which is divided into 5 categories according to the distribution of the population, variance, and standard deviation, and subject to changes in sub-~population and special cohort testing. Assuming that population is normal or quasi normal, the coefficient has been attributed according to the underlying thinking process of the test subject, the length of the brain processing being considered an indicator of the maturity of the brain. As discussed in section~\ref{ANN_s}, we are deploying a 5 categories model. 
Mathematically, figure~\ref{normal} shows how we define the five categories, with 0.5 to 1.5 seconds corresponding to the range of $[-3\sigma + \mu, -2\sigma + \mu]$, 1.5 to 2.5 seconds corresponding to the range of $[-2\sigma + \mu, -\sigma + \mu]$, 2.5 to 3.5 seconds corresponding to the range of $[-\sigma + \mu, \sigma + \mu]$, 3.5 to 4.5 seconds corresponding to the range of $[\sigma + \mu, 2\sigma + \mu]$, and 4.5 to 7 seconds corresponding to the range of $[2\sigma + \mu, 3\sigma + \mu]$ where $\mu$ and $\sigma$ are for mean and standard derivation.

The third part is the Biometric Type (BT) which can be divided into 4 categories assuming that is the average survey type between highest positive (Yes answers) emotion score and lowest negative (No answers) emotion score; or a coefficient can be attributed to the emotion in virtue of the social interest and communication value of showing average emotional reaction. For example in the categorical method: +1 is for Contempt or Disgust; +2 is for Anger or Fear; +3 is for Happiness or Sadness; +4 is for Surprise or Neutral. In the continuous method, a mapping is needed to convert and display the valence-arousal pair into the 4 categories system. A possible method is to divide the valence-arousal 2 dimensional plane into 4 regions representing the 4 categories as shown in right panel of figure~\ref{IWI1}.

As a result of having these scores calculated, the calculus of the final index (IWI) is greatly simplified by the following formula: 
$$
\textrm{IWI} = [(\textrm{RP} \times \textrm{T}) \times \textrm{TT} \times (\textrm{BT} \times \textrm{C})] 
$$
Therefore, the index can range in theory between 0 and 120. Maximum is $[6\times5\times4]$ assuming T=100\% and C=100\%. For practical reason, the proposed range in app is set between 20\% and 100\%.

\subsection{MIMAMO-NET for video emotion recognition}
MIMAMO Net, which is for abbreviation of Micro-Macro Motion Net, is the first to apply phase differences to valence-arousal estimation and achieves significant gains. As shown in figure~\ref{mimamo-net}, it consists of two streams of convolutional neural networks to model the micro-motion and a recurrent neural network to model the macro-motion. Here the micro-motion and macro-motion are defined as emotion-related facial motions in the micro and macro time scales. The MIMAMO Net samples a sequence of snippets as the inputs and returns the valence-arousal estimations for each snippet, where a snippet includes one RGB image and 13 grayscale images. During inference, the sequence length is 64, which corresponds to about 2 seconds (macro-scale), and the snippet length is 13, which corresponds to about 400 milliseconds (micro-scale). 

When the subject is presented with risk related questions of the SRTA, MIMAMO Net can analyze both the instantaneous facial movements in a very short time period and the consistent facial movements in a relatively long time period, which is superior to frame-based emotion recognition techniques. Based on the accurate estimations of valence-arousal scores and other inputs, our neural network can better analyze the risk tolerance and individual worthiness. 

\section{Related works and future research direction}

There are already many works on understanding the brain mechanisms and establishing a mathematical framework for brain-inspired machine learning systems~\cite{wang2015formal, wang2006layered, wang2020abstract, wang2016cognitive, wang2007theoretical}. But these works are still preliminary and stay at abstract and theoretical level and are not ready for direct real-life applications. 
However, it does not offer insight on how we should develop an artificial neural network model for the assessment of human characteristics such as Risk Profile, Thinking Type, Truthfulness, Leadership and Individual Worthiness while these are at the center of our model.

Our proposed model is taking a radically different approach by integrating well established neural network architecture~\cite{hassoun1995fundamentals, lecun2015deep, wasserman1993advanced, white1992artificial}, state-of-the-art emotion recognition model MIMAMO-Net~\cite{deng2019mimamo} and insights from neuropsychology~\cite{cloninger1986unified, cloninger1987systematic, cloninger1991tridimensional, cloninger1994temperament} to develop a system for the assessment of Individual Worthiness Index (IWI) and other parameters as discussed in section~\ref{IWI}. To the best of the authors' knowledge, this work is the first to propose a feasible approach to assess the risk profile of the test subjects based on how they react to a set of questions related to situational risk in a multidimensional personality questionnaire~\cite{cloninger1991tridimensional, cloninger1994temperament} with the aid of the latest technology in video emotion recognition MIMAMO-Net, and in a completely automated manner.

On the other hand, there are extensive studies on creditworthiness. The studies of econometric for creditworthiness assessment in credit market~\cite{broecker1990credit, meier2012time, bai2019banking} and the studies of targeting customers with a forecasting of their creditworthiness~\cite{helgesen2008targeting, siekelova2017receivables, dastoori2013credit} focus on applying conventional psychology and financial methods to the assessment of creditworthiness such as using Time discounting and FICO score to forecast creditworthiness of customers in~\cite{meier2012time}, without leveraging the latest advances in artificial intelligence. Our proposed model is inspired by traditional neuropsychology but also applies the most recent developments in neuromorphic intelligence to a novel machine learning framework, which is our main contribution. We are the first to design a labeled data-set using cognitive tags of 6 categories (the "question types" discussed in section~\ref{ANN_s}) to train our machine learning model. in traditional methods, after training, the machine learning model can be applied to data so that new unlabeled data can be presented to the model and a likely label can be estimated or predicted for the unseen new data. In our method, we are using our proprietary labeled dataset inspired by the TCI~\cite{cloninger1994temperament}.

Our model was built to establish the relationship between specific neural activities and financial behaviors using personal attributes captured live and real-time and delivering the Risk Tolerance score and Individual Worthiness index of the subject whether in the form of creditworthiness index for borrowing money or trustworthiness index for conduct risk management. There are research areas waiting to be improved, for example, the distribution model for the Thinking Type (TT) could become more accurate after a comprehensive and in-depth study on the brain processing time or average test user latency. One possible suggestion is that a skew normal distribution is more accurate for some cultures or ages. This work intends to define a new direction for future research in Neurofinance using brain-inspired artificial intelligence.

\section{Conclusion}
In this paper, we have reviewed the different aspects of the background theory of human personality testing by the acclaimed TCI of Cloninger and described how that has motivated the construction of a novel neuropsychological performance test leveraging state-of-the-art artificial neural network methods and video emotion recognition model MIMAMO-Net. We have discussed the architecture and details of our proposed risk profiling system, the structure of the ANN, the major and minor question types, the training process, the calculation, range, and assumptions of the parameters including the Individual Worthiness Index and finally the application of MIMAMO-Net. We mentioned that although there are extensive works on understanding the mechanisms of the brain and establishing a mathematical framework for brain-inspired neural network, these models stay at theoretical and abstract level which makes real-life application quasi impossible. We also discussed related works on assessment of creditworthiness in studies addressing econometric and credit market. While they focus on conventional psychology and financial market theories, we intend to implement solutions operating with artificial intelligence and emotion recognition. 

\section{Acknowledgment.}
The theory presented in this paper is based on two of our patents, namely, "Method for Neuropsychological Performance Test", Frederic Andre Jumelle and Yat Wan Lui, Hong Kong short-term patent grant certificate No. HK30003521A, and "Method and System for Neuropsychological Performance Test", Frederic Andre Jumelle and Yat Wan Lui, international patent application No. PCT/CN2019/095325.


%

%
%
%


\ifCLASSOPTIONcaptionsoff
  \newpage
\fi



%
%
%

\bibliographystyle{unsrt}
\bibliography{NIPS_Bib}

%








\end{document}